\documentclass[sn-nature]{sn-jnl}

\usepackage{graphicx}%
\usepackage{multirow}%
\usepackage{amsmath,amssymb,amsfonts}%
\usepackage{amsthm}%
\usepackage{mathrsfs}%
\usepackage[title]{appendix}%
\usepackage{xcolor}%
\usepackage{textcomp}%
\usepackage{manyfoot}%
\usepackage{booktabs}%
\usepackage{algorithm}%
\usepackage{algorithmicx}%
\usepackage{algpseudocode}%
\usepackage{listings}%
\usepackage{soul}%
\usepackage{makecell}%
\usepackage{booktabs,makecell,pifont}
\usepackage{pdfcomment}

\usepackage{geometry}
\newcommand{\cmark}{\ding{51}}
\newcommand{\xmark}{\ding{55}}

\theoremstyle{thmstyleone}%
\theoremstyle{thmstyletwo}%

\theoremstyle{thmstylethree}%

\begin{document}

\title[Accelerating High-Throughput Catalyst Screening by Direct Generation of Equilibrium Adsorption Structures] {Accelerating High-Throughput Catalyst Screening by Direct Generation of Equilibrium Adsorption Structures}


\author[1]{\fnm{Songze} \sur{Huo}}

\author*[1]{\fnm{Xiao-Ming} \sur{Cao}}\email{xmcao@sjtu.edu.cn}

\affil*[1]{\orgdiv{State Key Laboratory of Synergistic Chem-Bio Synthesis, School of Chemistry and Chemical Engineering}, \orgname{Shanghai Jiao Tong University}, \orgaddress{\street{800 Dongchuan Road}, \city{Shanghai}, \postcode{200240}, \state{Shanghai}, \country{China}}}


\abstract{The adsorption energy serves as a crucial descriptor for the large-scale screening of catalysts. Nevertheless, the limited distribution of training data for the extensively utilised machine learning interatomic potential (MLIP), predominantly sourced from near-equilibrium structures, results in unreliable adsorption structures and consequent adsorption energy predictions. In this context, we present DBCata, a deep generative model that integrates a periodic Brownian-bridge framework with an equivariant graph neural network to establish a low-dimensional transition manifold between unrelaxed and DFT-relaxed structures, without requiring explicit energy or force information. Upon training, DBCata effectively generates high-fidelity adsorption geometries, achieving an interatomic distance mean absolute error (DMAE) of 0.035 \text{\AA} on the Catalysis-Hub dataset, which is nearly three times superior to that of the current state-of-the-art machine learning potential models. Moreover, the corresponding DFT accuracy can be improved within 0.1 eV in 94\% of instances by identifying and refining anomalous predictions through a hybrid chemical-heuristic and self-supervised outlier detection approach. We demonstrate that the remarkable performance of DBCata facilitates accelerated high-throughput computational screening for efficient alloy catalysts in the oxygen reduction reaction, highlighting the potential of DBCata as a powerful tool for catalyst design and optimisation.}

\maketitle
 
\section{Introduction}
Adsorption plays a central role in heterogeneous catalytic reactions. It provides essential insights into the interactions between the catalyst surface and reactants, thereby constituting a critical step in catalyst design \cite{Heterogeneous_Thomas_2014}. The adsorption energy serves as the most vital numerical parameter in the adsorption process and is a significant descriptor of catalytic activity\cite{Cata_Norskov_2009}. Accurate calculation of adsorption energy is intrinsically linked to the precise prediction of the adsorption structure, which requires high-precision geometry relaxation. Currently, the most commonly used method for predicting adsorption structures is based on first-principles calculations, such as density functional theory (DFT) \cite{DFT_Hohenberg_1964, DFT_Kohn_1965}. However, the computational complexity of the DFT method is $O(n^3)$, where $n$ is the total number of electrons in the system \cite{AdsorbML_Ulissi_2023}. Additionally, DFT requires self-consistent field (SCF) iterations to ascertain the energy and forces of the system, followed by repeated structural optimisations to determine the adsorption structure, which demands significant computational resources. Consequently, predicting adsorption structures remains a major bottleneck in the rational design of catalysts, underscoring the necessity to develop efficient methods to expedite this process.

Recent surges in computational power have accelerated the development of deep learning algorithms, facilitating the creation of large-scale public datasets within the catalysis community \cite{CatHub_Thomas_2019, OC20_2021, OC22_2023, aqcat25}. Deep learning models have demonstrated remarkable success in a variety of supervised learning tasks in catalysis, such as directly predicting adsorption energies from hand-crafted or learned descriptors without explicitly performing geometry relaxation \cite{ACE-GCN_2022, GAMENET_2023, DOTA_2025, ads_2018, graphads_2023, CatBERTa_2023, GMAE_Chen_2025}. Nevertheless, most existing approaches are tailored to specific surfaces and adsorbates, thereby lacking generalisability. These models typically frame adsorption energy prediction as a regression or classification problem and rely on static chemical descriptors or string-based language tokens as inputs, which renders the reliability of the model structurally dependent. Crucially, adsorption energies are highly sensitive to the detailed local adsorption geometry, such as the adsorption site, molecular orientation, and coordination environment, so faithfully capturing the structure-energy relationship requires methods that can explicitly reason about and generate plausible adsorption configurations. To move beyond adsorption energy prediction from static representations, recent works have started to incorporate machine learning into workflows that directly explore the configurational space of adsorbate-surface geometries and identify energetically low-lying adsorption states \cite{globalopt_2023, adsorbdiff_2024, ActiveLearning_Ulissi_2024}. Jung et al.\ developed a constrained minima-hopping workflow combined with on-the-fly DFT  evaluations and Gaussian approximation potential (GAP) training to efficiently identify unbiased adsorption minima on catalytic surfaces \cite{globalopt_2023}. Collectively, these methods have significantly advanced the use of machine learning for accelerating the search and optimisation of adsorption configurations. However, they still depend on repeated evaluations of high-fidelity DFT energies and forces within iterative optimisation loops, which dominate the computational cost and ultimately limit their efficiency and scalability for very large design spaces. This strong dependence on inner-loop DFT evaluations has motivated efforts to replace DFT with learnt surrogate models during structure relaxation.

Building on this idea, a more radical alternative is to remove DFT from the relaxation loop by using machine learning interatomic potential (MLIP) models that directly predict the energies and forces of atomic configurations \cite{EGNN_Satorras_2021, PaiNN_Schutt_2021, mlff_2021_schutt, mace_2022, eqv2_Liao_2023}. Once trained, such models can be used as drop-in replacements for DFT in geometry optimisation to obtain relaxed adsorption structures at a fraction of the cost \cite{Perspective_Xu_2021, FF_2022_Google}. In parallel, a growing number of general-purpose, pretrained ML potentials have been proposed for catalysis, where a single model is reused across many surfaces and adsorbates. For example, AdsorbML employed a library of pretrained graph neural network potentials to relax and rank large numbers of adsorbate-surface configurations \cite{AdsorbML_Ulissi_2023}, illustrating how such foundational models can be leveraged in practical screening workflows. However, the effectiveness of these MLIP-based strategies is fundamentally limited by the quality and coverage of the underlying training data. Since geometry optimisation generally starts from a structure offsetting a local minimum, the entire optimisation process is likely to explore the global potential energy surface (PES). This implies that training a robust MLIP requires a sufficiently large dataset that adequately samples the complex high-dimensional PES, posing significant challenges and placing substantial demands on both model size and computational resources during training. The accuracy of the predicted energies and forces is highly sensitive to the coverage and information content of the training set. The absence of non-equilibrium structures in the datasets that are mainly composed of optimised structures and near-equilibrium molecular dynamics simulations can exacerbate inaccurate energy and forces of the unrelaxed structure, thereby leading to the deviated geometry optimisation trajectory from the correct PES and the unreliable adsorption structures \cite{data_2024}. These limitations constrain the broader adoption of MLIP-based workflows for accelerating catalyst design and motivate alternative approaches that directly learn the transformation from initial to relaxed adsorption structures without explicitly fitting the full PES.

An alternative viewpoint is to treat adsorption geometry prediction as a direct sampling problem on the PES. Given an initial, off-equilibrium adsorbate-surface configuration, the optimised adsorption structure is generated by sampling from the conditional distribution of relaxed, low-energy structures consistent with the underlying thermodynamics. Recent advances in deep generative modelling have demonstrated that such high-dimensional equilibrium distributions over molecular conformations and material structures can be learnt and efficiently sampled in practice, using diffusion models, normalising flows, and score-based methods to approximate chemical data distributions and generate physically plausible 3D configurations \cite{CDVAE_2021, OA_Duan_2023, react-ot_2025, mattergen_2025}. In particular, diffusion-based models have been successfully applied to molecular conformation generation and crystal or surface structure design, where they learn to map noise to ensembles of equilibrium structures while respecting geometric and symmetry constraints through equivariant neural networks \cite{geodiff_2022, edm_2022, con-cdvae_2024, surfdiff_2024}. Complementary flow-based approaches, such as continuous normalising flows and flow matching, offer faster and often deterministic sampling by learning velocity fields that transport one probability distribution into another \cite{NormalizingFlows_2021, fm_2022, rflow_2022}, but they are typically defined between unconditioned distributions and can become brittle when asked to transform a specific initial configuration into a chemically valid relaxed state under large distribution shifts or limited data \cite{db_or_fm_2025}. Diffusion bridge models address this limitation by constructing stochastic processes that are explicitly constrained by both the initial and terminal distributions, thereby providing an intrinsic formulation of conditional structure generation and equilibrium sampling between off-equilibrium inputs and relaxed targets \cite{DiffSBDD_2023, BBDM_Li_2023, ddbm_2023}. 

Inspired by the abovementioned advantages of the Brownian Bridge model, we introduce a novel DBCata (Diffusion Bridge Model for Catalysts) framework, directly encoding the transformation from initial to relaxed adsorption structures by the Brownian Bridge, to rapidly and accurately determine adsorption structures and obtain adsorption energies. DBCata is designed to accelerate the rational design of catalysts by integrating a generative model with an equivariant graph neural network, enabling the generation of stable adsorption structures for adsorbates on surfaces in less than 1 second, which is approximately 10,000 times faster than traditional DFT calculations that typically require hours or even days for structure optimisation. Unlike existing MLIP models, DBCata does not require large datasets that span the entire PES. Instead, it leverages a self-supervised learning method to construct a low-dimensional manifold that captures the transition between initial and relaxed structures. This allows DBCata to be trained on a relatively small dataset containing only paired structures, without the need for energies or forces, thereby accelerating the convergence of the model and significantly reducing resource consumption. Using the Catalysis-Hub dataset as a benchmark, we demonstrate that DBCata outperforms the state-of-the-art machine learning potential model, UMA-OC20 \cite{uma}, from the Open Catalyst Project and FAIRChem \cite{fairchem2025}, both in terms of speed and accuracy. Furthermore, we introduce an innovative outlier detection method that employs controllable stochastic generation and chemical heuristics to identify and refine outlier structures. It shows that the robustness and high efficiency of DBCata enable accelerating high-throughput computational screening for alloy catalysts in the oxygen reduction reaction (ORR). The results showcase how DBCata can dramatically reduce the computational cost associated with DFT or MLIP relaxation, underscoring its potential as a powerful tool for catalyst design and optimisation.

\section{Results}

\subsection{Model overview} \label{model overview}

\begin{figure}[h]
    \centering
    \includegraphics[width=1.0\textwidth]{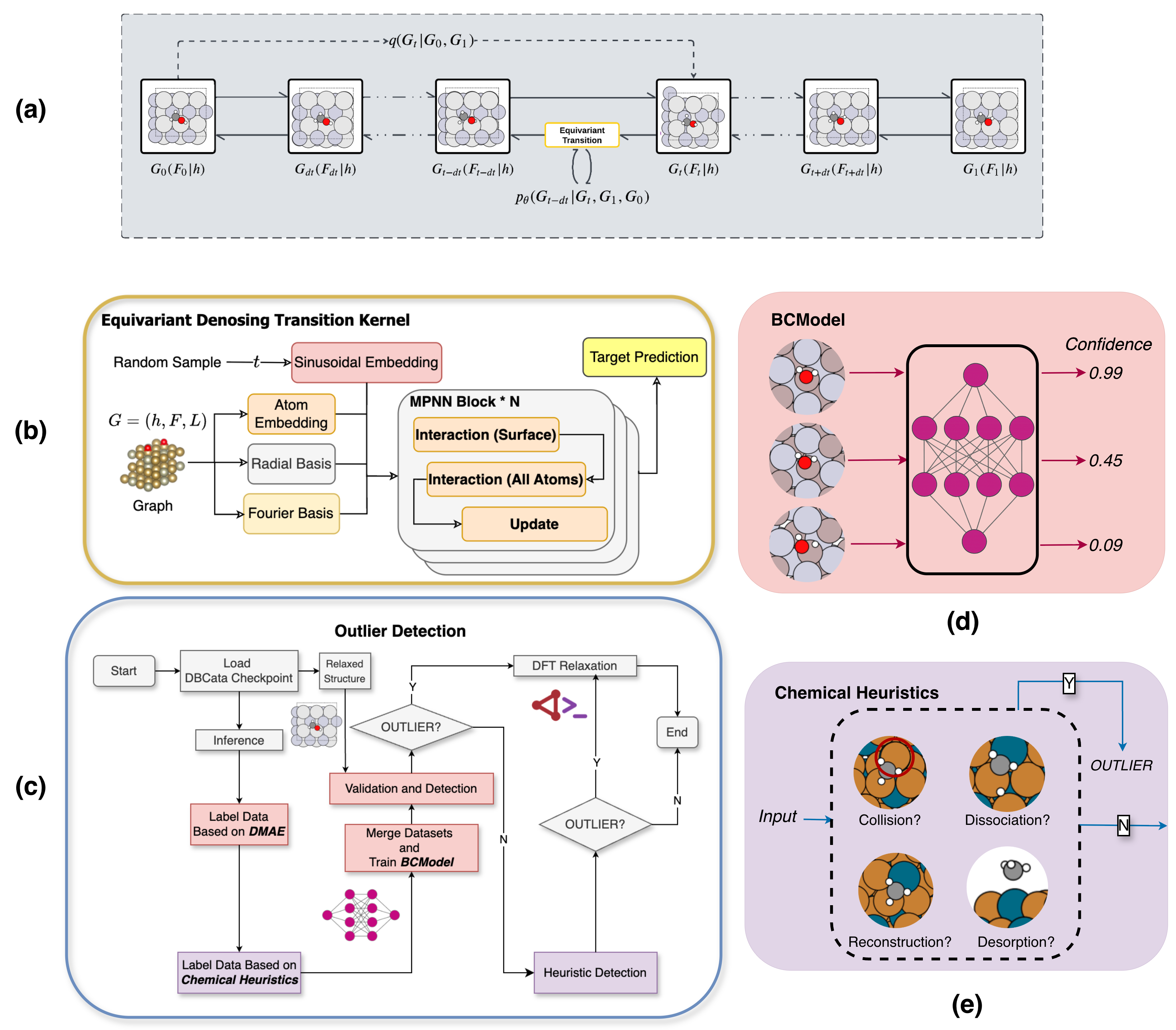}
    \caption{\textbf{Architecture of DBCata}. (a) Forward diffusion process and backward denoising process. (b) Equivariant graph neural network for predicting denoising objects. \textbf{Workflow of outlier detection.} (c) Workflow of labelling, training, and detecting. (d) The classifier model with the output of the confidence level of the detected structure. (e) Identification of outliers based on chemical heuristics.}
    \label{overview}
\end{figure}

In contrast to mainstream generative models such as VAEs \cite{vae_2013} and diffusion probabilistic models \cite{ddpm_ho_2020, sgm_song_2020}, which establish a mapping from a simple prior distribution (e.g., Gaussian) to the data distribution, DBCata is built on a Brownian-bridge formulation. During the training phase, it is conditioned on both the initial and relaxed structures, employing interpolation to learn a neural network mapping between their probability distributions. This approach circumvents the need to model the entire complex PES. After training, the model can generate relaxed structures directly on the learnt virtual low-dimensional manifold by providing an initial structure. The overall architecture of the model is illustrated in \textbf{Fig.\ref{overview}}. During training, which corresponds to the diffusion process, Gaussian noise with a specified variance was added to a linear interpolation between the initial and relaxed structures to simulate the stochastic path of a Brownian bridge. By introducing noise with varying variance during training, the model covers a sufficiently broad range of paths. Similar to the other generative models, maximising the Evidence Lower Bound (ELBO) is the training objective. During the generation phase, an Ordinary Differential Equation (ODE) approach was implemented for deterministic sampling, ensuring that the generated structures are unique and invariant upon repeated sampling. Detailed information about DBCata is available in Section \ref{DBCata Method}.

To accurately model interatomic interactions, the Polarizable Atom Interaction Neural Network (PaiNN) \cite{PaiNN_Schutt_2021} architecture was employed as the transition kernel for DBCata, leveraging its equivariant properties to ensure compliance with physical symmetries. Given that adsorption takes place on periodic slabs, it is crucial to consider the periodic boundary condition (PBC). To ensure periodicity of interatomic interactions, a periodic Brownian bridge model was developed (detailed in Section \ref{DBCata Method}), which utilises a multi-graph approach to identify the neighbours of each atom within the cutoff radius across multiple periodic images. Additionally, a nearest-atom interpolation method was introduced to guarantee that the transition probabilities during the Brownian bridge process inherently satisfy periodic boundary conditions. 

To mitigate rare but noticeable outliers produced by DBCata, a hybrid outlier-detection scheme was implemented, which combines chemical heuristics with a binary anomaly classifier (\textbf{Fig.\ref{overview}c-e}). Initially, we employed the pretrained DBCata model to generate numerous relaxed structures across varying noise levels. The Distance Mean Absolute Error (DMAE), which measures the error between the model-generated structures and the DFT-relaxed structures, was subsequently employed as a metric to efficiently evaluate the reliability of the pretrained DBCata model. The smaller the DMAE value is, the closer the prediction is to the DFT-relaxed structure. Meanwhile, those with DMAE values exceeding a predefined threshold were identified as potential outliers. Additionally, chemical heuristics were employed to detect chemically unreasonable structures, such as those with interatomic distances significantly shorter than the sum of their covalent radii. Next, a binary anomaly detection classifier was trained to automate the identification of outliers. Finally, these outlier structures were refined through DFT relaxation.

\subsection{Accurate structure prediction and enhanced refinement with outlier detection} 

Catalysis-Hub is a publicly available catalytic dataset containing approximately 80,000 trajectories of relaxed metal and alloy surfaces computed using DFT with the BEEF-vdW functional \cite{beef_2012}. From this dataset, we successfully extracted 76,737 pairs of relaxed structures and their corresponding initial structures. The included adsorbate species are H, C, CH, CH$_2$, CH$_3$, N, NH, O, OH, H$_2$O, S, and SH. Subsequently, we cleaned the original dataset by removing structures that showed no changes or exhibited excessive changes, which resulted from optimisation errors in structural optimisation by DFT. The cleaned data were randomly split into training and test sets in an 8:2 ratio (58,943:14,728) for training and evaluating DBCata. Due to the model's computational efficiency and relatively small number of parameters, training can be completed within 12 hours using one Nvidia RTX 4090 GPU card. DBCata achieved a DMAE of 0.0352 \text{\AA} on the test dataset, outperforming the 0.1133 \text{\AA} DMAE of the optimised structures using the pretrained universal MLIP UMA-OC20. DBCata generates relaxed structures in less than 1 second per structure, approximately $4\times10^{3}$ times faster than conventional DFT relaxation and 10 times faster than UMA-OC20 relaxation (without parallelisation). Performance can be further improved via GPU parallelisation. Complete generation-efficiency results and comparisons are provided in Supplementary Table 1.

DBCata also exhibits the generality across different adsorbates and exposed facets. The dataset for DBCata training covered different adsorbates and exposed facets. Supplementary Table 2 shows that the DMAE values of different adsorbates range from 0.0268 to 0.0667\text{\AA}, consistently smaller than that of the optimised structures by UMA-OC20. This indicates that DBCata could be applied to various adsorbate species. Moreover, similar DMAE values of 0.0341 and 0.0361\text{\AA} were yielded for fcc(111) and fcc(110), respectively. Collectively, these results indicate that DBCata effectively captures the inherently structural features of the adsorption processes, thereby being possibly applied to different adsorbates and exposed facets.

\begin{figure}[h]
    \centering
    \includegraphics[width=1.0\textwidth]{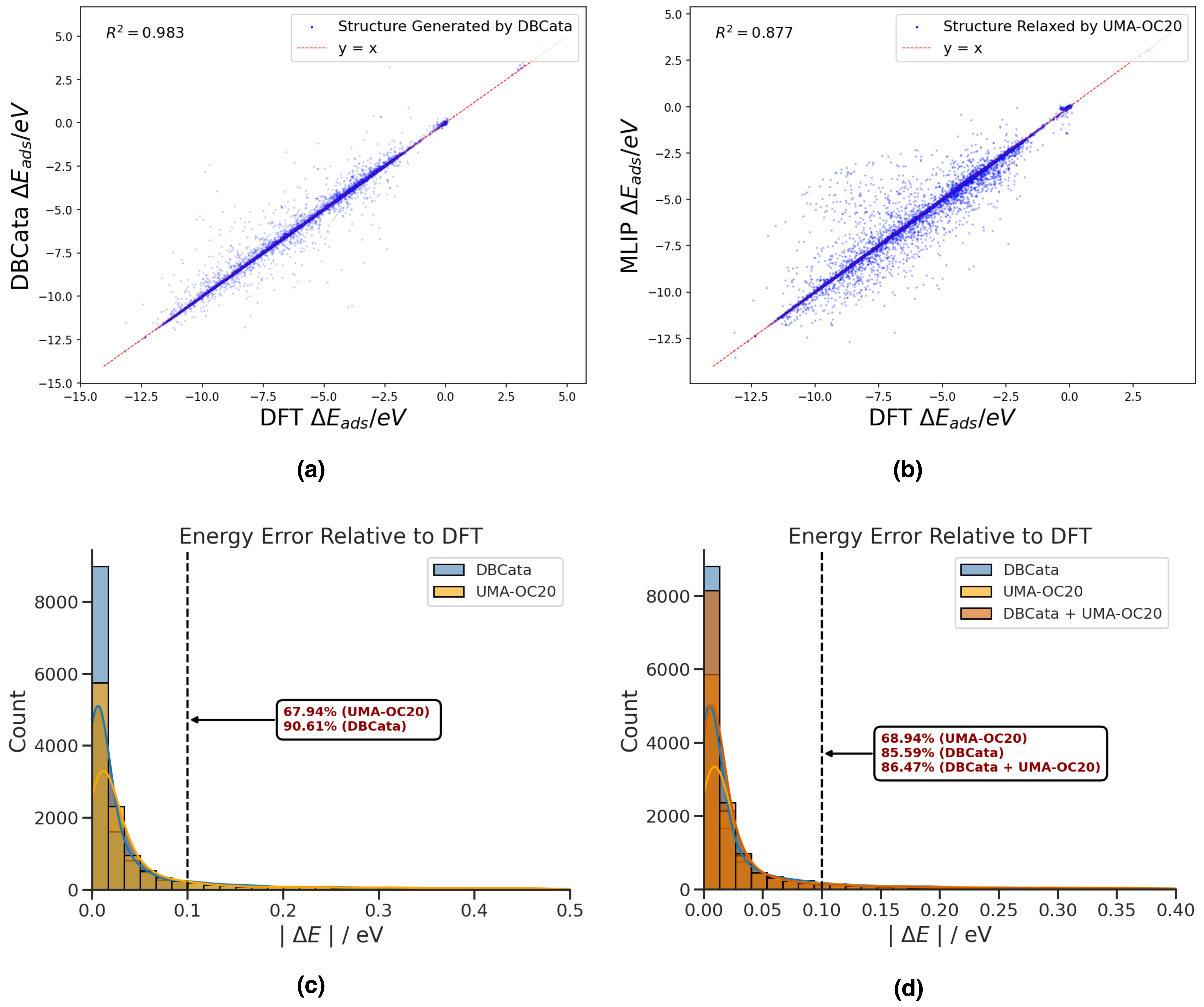}
    \caption{(a-b) Parity plots comparing adsorption energies derived from DFT-relaxed structures versus model-predicted structures. The data correspond to the Catalysis-Hub test dataset, with all energies calculated using DFT. The red dashed line indicates perfect parity. (c) Distribution of absolute energy errors ($|\Delta E|$) relative to DFT-relaxed structures, evaluated using DFT calculations. The comparison includes DBCata-generated structures (blue) and UMA-OC20 relaxed structures (yellow). (d) Distribution of energy errors evaluated using the UMA-OC20 potential relative to DFT-relaxed structures. This includes DBCata-generated structures (blue), UMA-OC20 relaxed structures (yellow), and DBCata structures post-relaxed by UMA-OC20 (brown). The region within the black dashed lines denotes predictions within an accuracy threshold of 0.1 eV.} 
    \label{cathub_results}
\end{figure}

Despite the high efficiency, the DMAE metric may sometimes not fully reflect the accuracy of predicted structures due to the symmetry of the adsorption configuration. For example, for the tilt adsorption of OH species atop the surface, various O-H bond orientations could correspond to a series of energetically equivalent adsorption structures on a specific surface. However, each training data point contains only one benchmark adsorption orientation. Thus, even if the predicted adsorption structure is identical to the benchmark DFT-relaxed one, the different bond orientation might also result in a spuriously elevated DMAE value (Supplementary Figure 3). Therefore, the adsorption energy based on the DFT-relaxed structure was also employed as a more robust metric to confirm the performance of DBCata in the adsorption energy prediction. Specifically, DFT single-point calculations using the revised Perdew-Burke-Ernzerhof (RPBE) \cite{rpbe} functional were performed on DBCata-generated structures in the test dataset to obtain adsorption energies, which could also be compared with the RPBE results using the UMA-OC20-relaxed structures. All adsorption energies were computed referencing gas-phase adsorbates, as detailed in Section \ref{Computational details}. As shown in \textbf{Fig.\ref{cathub_results}a-b}, the calculated adsorption energies based on DBCata-generated structures are significantly closer to the DFT-relaxed results than those based on UMA-OC20-relaxed structures. Using an energy threshold of 0.1 eV, DBCata demonstrated a prediction accuracy success rate of around 0.91, outperforming 0.68 for UMA-OC20 relaxed (\textbf{Fig.\ref{cathub_results}c}). These results indicate that DBCata exhibits exceptional performance in accurately predicting fine structural details. 

\begin{figure}[!htbp]
    \centering
    \includegraphics[width=1.0\textwidth]{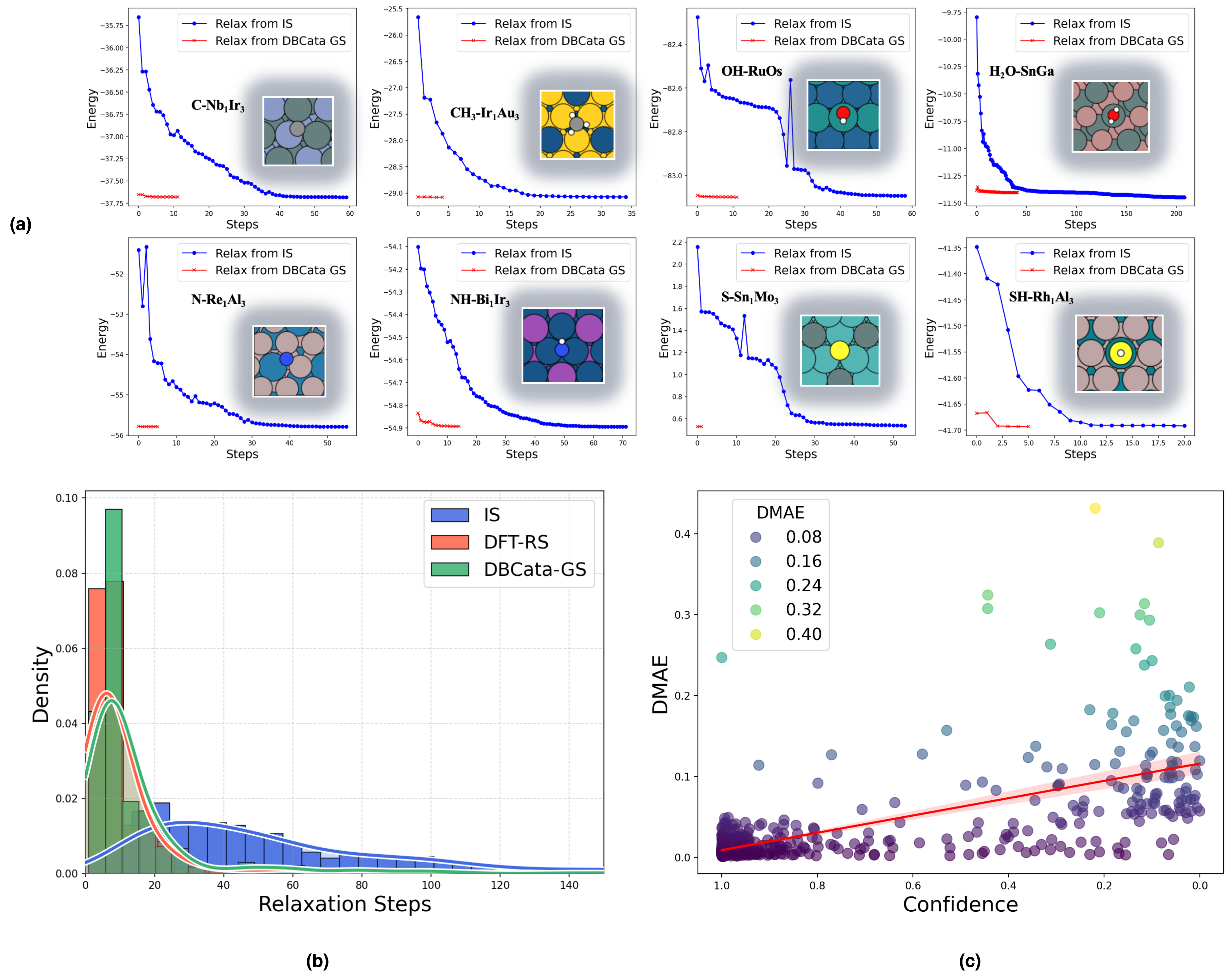}
    \caption{(a) Comparison of ionic steps required for DFT relaxation starting from initial structure (IS) versus DBCata-generated structures (DBCata-GS) for species CH$_x$, OH$_x$, NH$_x$, and SH$_x$. (b) Distribution of ionic steps required to reach tight convergence during additional DFT relaxation for initial structures (IS), DBCata-generated structures (GS), and DFT-relaxed structures (RS). (c) Correlation between classifier confidence and DMAE for generated structures; points farther to the right correspond to lower confidence (i.e., higher uncertainty).}
    \label{relaxation}
\end{figure}

To further enhance the accuracy of the predicted structures, post-relaxation DFT calculations can be performed on DBCata-generated geometries using a tighter force convergence criterion. Compared with direct DFT relaxation from the initial structure (IS), this procedure substantially reduces the number of required ionic steps (\textbf{Fig.\ref{relaxation}b}). Representative results for a subset of 500 structures (comprising 100 H, 100 CH$_x$, 100 NH$_x$, 100 OH$_x$, and 100 SH$_x$ adsorbates) selected from the test set are shown in \textbf{Fig.\ref{relaxation}a}. On average, direct DFT relaxation from the initial configurations required 65 ionic steps, whereas starting from DBCata-generated structures required only 15 steps. Furthermore, when re-relaxing the structures from the BEEF-vdw-relaxed structures in the Catalysis-Hub test set to the same convergence criteria using the RPBE functional, an average of about 13 additional ionic steps was still required to reach tight convergence. Notably, DBCata accelerates the search for the adsorption structure. 

While most structures generated by the DBCata model align well with chemical intuition and exhibit high accuracy, a small number of outliers deviate from the true configurations. To address this issue, we developed the hybrid outlier detection approach mentioned in \ref{model overview}. To assess the efficacy of our proposed approach, we conducted structure generation on the Catalysis-Hub test dataset and compared the calculated DMAE values against predictions from our anomaly detection classifier. The classifier outputs confidence scores indicating uncertainty levels in generated structures. As illustrated in \textbf{Fig.\ref{relaxation}c}, structures with lower classifier confidence (high uncertainty) correlate strongly with elevated DMAE values, confirming the effectiveness of the classifier. Subsequently, we performed DFT relaxation exclusively on the structures flagged as outliers by high uncertainty or chemical heuristics violations. The outcomes related to computational efficiency and prediction accuracy are summarised in Table \ref{tab1}. Employing an energy threshold of 0.1 eV, the accuracy of DBCata with integrated outlier detection improved to 0.94, surpassing the accuracy of 0.91 achieved without outlier refinement.

\begin{table}[h]
    \caption{Comparison of accuracy and computational efficiency: Direct DFT relaxation, DBCata alone, and DBCata with outlier detection (OD) and subsequent DFT relaxation.}\label{tab1}
    \small
    \begin{tabular}{@{}llll@{}}
    \toprule
    Method & Num of Ionic Steps  & Accuracy Rate & DMAE\\
    \midrule
    DFT             &  $\sim1,000,000$    &   /   &     /      \\
    DBCata        &  /        &  0.91 &   0.0352   \\
    DBCata + OD   &  $30,108$     &  0.94 &   0.0281   \\
    \botrule
    \end{tabular}
    \end{table}

\subsection{Validating the near-equilibrium nature of relaxed structures on the PES}

\begin{figure}[h]
    \centering
    \includegraphics[width=1.0\textwidth]{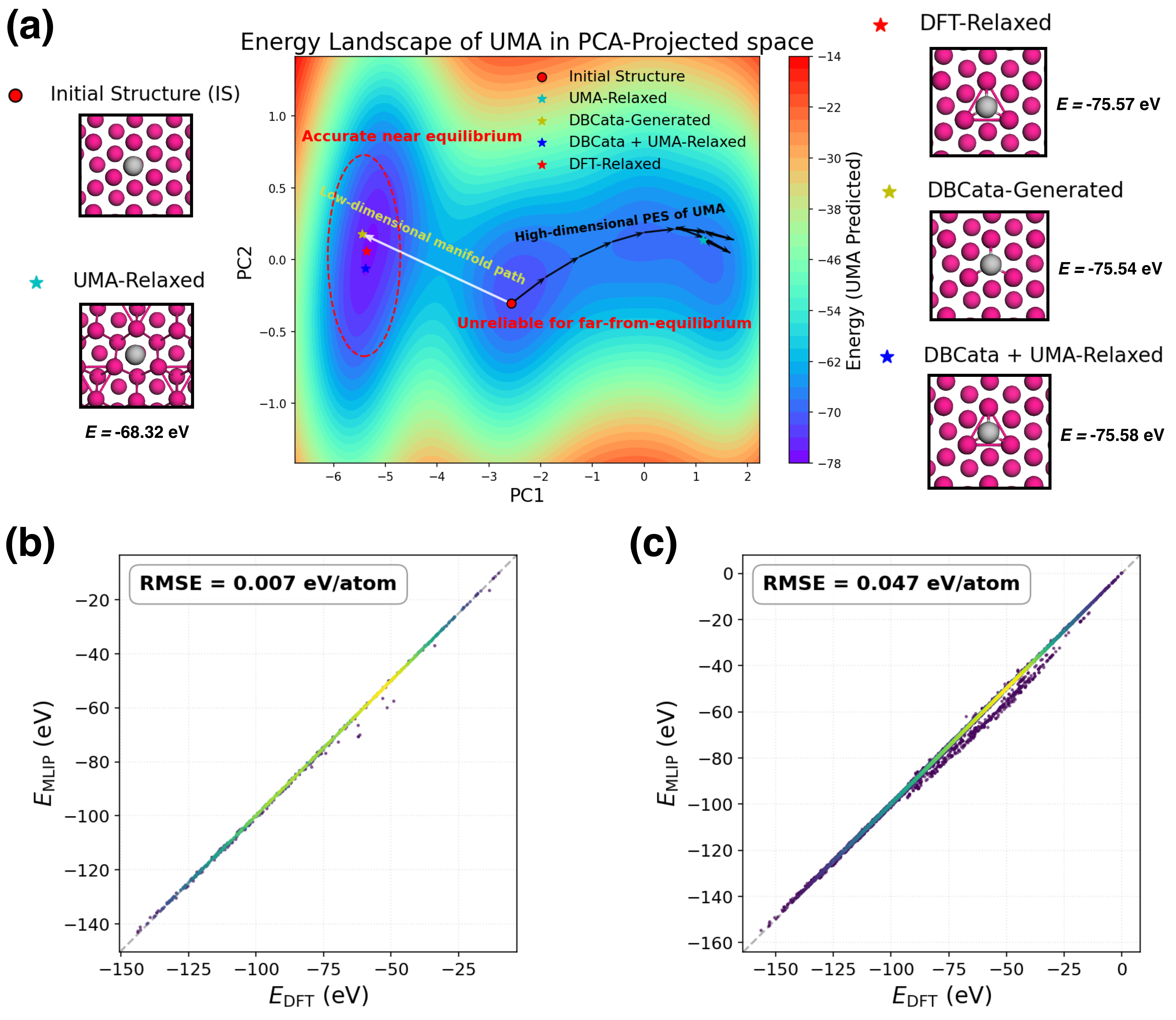}
    \caption{(a) Energy landscape for C adsorption at the fcc-hollow site on Y(111). The red dot, the red star, the yellow star, and the blue star mark the initial structure, the DFT-relaxed structure, the DBCata-relaxed structure, and the UMA-OC20-relaxed structure, respectively. The landscape was computed with UMA-OC20 and projected onto the plane spanned by the first two principal components (PC1, PC2) for visualisation. Solid black arrows indicate the MLIP optimisation trajectory, and dashed arrows indicate the DBCata generative path on the low-dimensional manifold. Filled energy contours are smoothed using kernel ridge regression with a radial basis function (RBF) kernel. (b) Energy evaluation of UMA-relaxed structures (5081 near-equilibrium structures). (c) Energy evaluation of UMA-relaxed structures (9647 far-from-equilibrium structures).}
    \label{landscape}
\end{figure}

Despite its robustness, the DFT adsorption energy metric is expensive for evaluating the model's performance. The MLIP-calculated adsorption energy is more desirable as the model metric. Moreover, it is expected that the refinement of the DBCata-generated adsorption structures with high uncertainty or chemical heuristics violations could be post-relaxed using MLIP to achieve the accurate adsorption energy. Therefore, we probed why MLIP optimisation results in poor adsorption energy data and deviates from the equilibrium state. A geometric criterion based on the DMAE between the initial and DFT-relaxed structures was applied to 14,728 test-set structures, approximately classifying samples with DMAE $> 0.1$ \text{\AA} as far-from-equilibrium and those with DMAE $< 0.1$ \text{\AA} as near-equilibrium. RPBE single-point calculations were then performed on UMA-relaxed structures to quantify the root-mean-square error (RMSE) of forces and energies predicted by the UMA model. The force RMSE for the far-from-equilibrium set (9,647 samples) is substantially higher than that for the near-equilibrium set (5,081 samples). Notably, Supplementary Figure 4 shows that forces on some surface atoms cluster near zero yet deviate strongly from the parity line, suggesting that MLIP structural optimisation becomes trapped in incorrect local minima. The energy predictions exhibit a consistent trend, with RMSE of 46.52 meV/atom and 6.99 meV/atom for the far-from-equilibrium and near-equilibrium subsets, respectively (\textbf{Fig.\ref{landscape}b-c}). Together, these results indicate that the MLIP is significantly less accurate for structures far from equilibrium, which can in turn induce spurious optimisation trajectories and degrade adsorption-energy estimates.

Therefore, a near-equilibrium state around a local minimum of the PES, where the MLIP is well described and its predictions of energies and forces are likely to be sufficiently accurate, could be generated by DBCata. Thus, we can use the MLIP for energy evaluations to test the accuracy of the DBCata-generated structures or perform a post-relaxation step to ensure that the generated structures are closer to this local minimum. As an illustration, \textbf{Fig.\ref{landscape}a} visualises the energy landscape of C adsorption at the fcc site on Y(111). The DBCata-generated structure and the DFT-relaxed structure lie very close to the local minimum of the UMA-PES and fall within the near-equilibrium region, whereas the initial structure and the UMA-relaxed structure are farther from equilibrium. Therefore, MLIP evaluations of DBCata outputs for the test dataset also show that DBCata-generated structures outperform UMA-OC20 (\textbf{Fig.\ref{cathub_results}d}), which is consistent with the DFT evaluations (\textbf{Fig.\ref{cathub_results}c}). Moreover, after post-relaxation with UMA-OC20, the prediction accuracy could be further improved. These consistencies further demonstrate that DBCata generates accurate structures and that the MLIP successfully fits the neighbourhood of the equilibrium state.

These results indicate that the structures generated by DBCata are very close to the local minimum of both DFT PES and MLIP PES, which can significantly reduce the computational cost of DFT relaxation or MLIP relaxation. Moreover, the combination of DBCata and MLIP could efficiently calculate the adsorption energies, thereby accelerating the process of catalyst design. 

\subsection{Accelerating catalyst design via automatic generation of adsorption structures}

Owing to their excellent properties, such as high catalytic activity and economic viability, alloy materials have gained extensive attention in various catalytic reactions \cite{alloys_nakaya_2022}. However, identifying alloys with specific compositions and structures that can maximise catalytic performance remains a significant challenge. In this study, we use the oxygen reduction reaction (ORR) as an illustrative case to demonstrate the capability of DBCata in accelerating high-throughput computational screening for efficient alloy catalysts.

Leveraging the end-to-end generation capabilities of DBCata, we directly employ computational tools such as Pymatgen \cite{pymatgen_Ong_2013} and CatKit \cite{catkit_thomas_2019} to generate ISs for the model input. These tools allow for the automatic creation of ISs with a fixed rule. Subsequently, DBCata enables the automated generation of relaxed adsorption configurations. Specifically, our study encompasses ORR investigations on 1,760 face-centred cubic (fcc) binary alloy systems, comprising 3d, 4d, and 5d transition metals and all 13 main-group metals (Al, Ga, In, Tl, and Nh). These alloys were explored with varying stoichiometric ratios (0.25, 0.5, and 0.75) and surface terminations (111 and 110). 

To identify promising binary alloy catalysts for ORR, stable structures with adsorbed O and OH species on all investigated surfaces were required. We systematically sampled all possible adsorption sites on each alloy surface, positioning adsorption molecules accordingly (details in Section \ref{IS}). In total, we initialised 15,911 initial adsorption structures spanning various catalyst compositions, surface terminations, and adsorption sites. With DBCata, rapid generation of numerous relaxed adsorption structures is achieved. Remarkably, structural relaxation using DBCata required merely 5 minutes, substantially outperforming traditional high-throughput DFT relaxations. The energetically favourable adsorption site and the corresponding adsorption energy of each surface were chosen for ORR catalyst screening. 

\begin{figure}[h]
    \centering
    \includegraphics[width=1.0\textwidth]{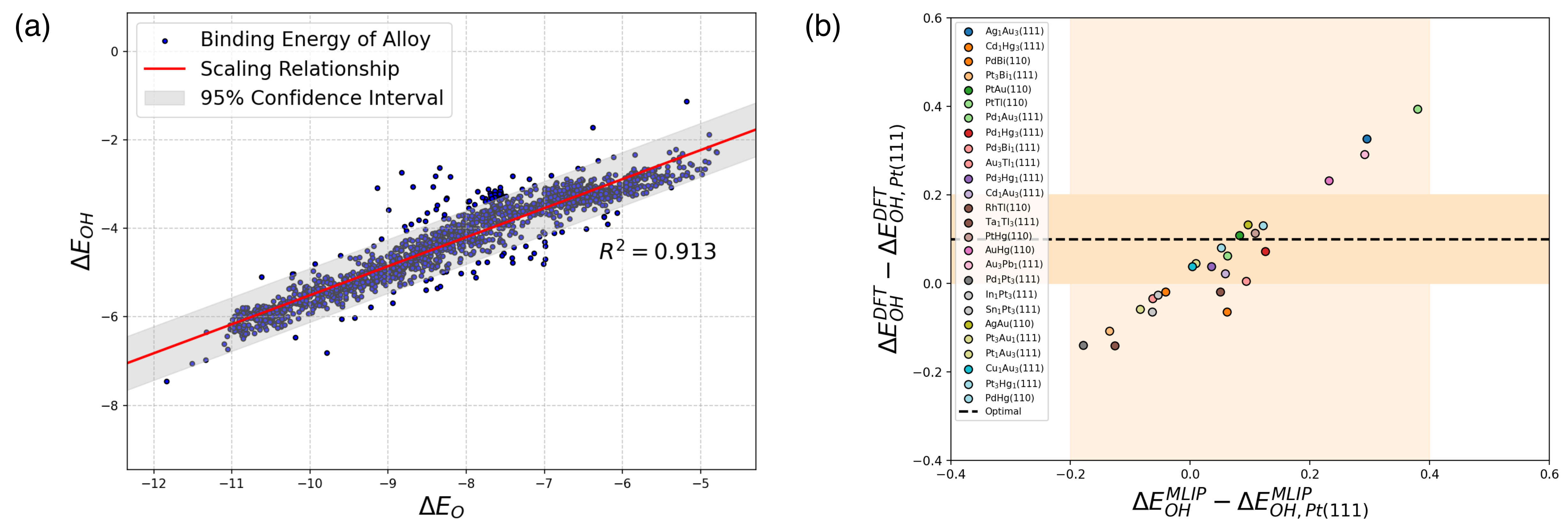}
    \caption{Calculation results of candidate catalysts. (a) Scaling relationship between OH and O adsorption energies derived from structures generated by DBCata. (b) Identification of promising candidates with slightly weaker adsorption energies than the reference Pt(111) surface; candidates falling outside the feasible range for DFT verification are excluded.}
    \label{optimal}
\end{figure}

Substantial experimental and theoretical evidence indicates that the adsorption energies of O and OH species are critical descriptors of ORR activity. Ideal catalysts exhibit slightly weaker O and OH adsorption (by approximately 0.2 eV and 0.1 eV, respectively) relative to Pt(111) \cite{orr_norskov_2018, ptorr_wei_2015}, which facilitates OH desorption and promotes the formation of OOH intermediates. In this work, we employed MLIP to efficiently compute the adsorption energies of O and OH on all alloy surfaces using the adsorption structures generated by DBCata, and thereby identifying the most promising candidates from these MLIP predictions. As illustrated in \textbf{Fig.~\ref{optimal}a}, the results reproduce the expected scaling relationship between the adsorption energies of O and OH, in agreement with established theoretical expectations \cite{scaling_2007, scaling_2016}. We emphasise that the generative model trained to predict accurate adsorption structures, together with the MLIP-based energy evaluation, is intended solely for preliminary high-throughput catalyst screening. By combining our outlier detection model with chemical heuristics and refinement, we identified 34 alloy surfaces whose optimal OH adsorption energy across all sites is approximately 0.1 eV weaker than that on Pt(111) (i.e., those satisfying $-0.2 \, \text{eV} < \Delta E_{OH}^{\text{MLIP}} - \Delta E_{OH, \text{Pt(111)}}^{\text{MLIP}} < 0.4 \, \text{eV}$) as potentially active candidates for further investigation. For this reduced set, post-relaxation DFT calculations were then performed to validate the MLIP-predicted adsorption energies and to further assess the activity and stability of the selected catalysts, as shown in \textbf{Fig.~\ref{optimal}b}. The overlapping yellow-shaded region corresponds to the alloy surfaces identified by this screening protocol as the most promising ORR catalyst candidates. 

Among the screened candidates, Pt- and Pd-based alloys exhibit the most promising theoretical characteristics, in line with experimental evidence for their high activity and stability \cite{orr_norskov_2018, ORR_Pd}. In particular, PtAu(110) emerges as a top candidate in our screening, a result consistent with reports that Au alloying enhances Pt stability against dissolution while simultaneously optimizing oxygen binding energetics \cite{ORR_PtAu, ORR_PtAlloy}. Although Pd-based catalysts often perform poorly in acidic media, our predictions for Pd-Au alloys agree with experimental studies identifying them as exceptional electrocatalysts in alkaline media, where they display superior kinetics and durability relative to pure Pd \cite{ORR_Pd, ORR_Pd_alkaline}. The Pt-Hg and Pd-Hg systems are predicted to sit near the apex of the ORR activity volcano, consistent with the d-band center model \cite{ORR_dband}, where the increased filling of antibonding states leads to weakened OH adsorption. In practice, however, these alloys are better viewed as model catalysts for the 2e$^-$ ORR to H$_2$O$_2$ \cite{2eORR}. Their tendency to form liquid amalgams precludes stable, well-defined facets, and the extreme toxicity and environmental mobility of mercury make them unsuitable for sustainable 4e$^-$ ORR technologies \cite{Hg_toxicity}. The predicted strong binding on Ta- and V-based alloys reflects their intrinsic oxophilicity, placing them on the strong-binding side of the ORR volcano. Consistent with this, experiments indicate that these materials undergo rapid surface passivation into insulating oxides rather than maintaining metallic catalytic activity in acidic media \cite{ORR_Ta}. Alloys such as Ag$_1$Au$_3$ and Pd$_1$Au$_3$ exhibit highly positive adsorption energy shifts, indicating binding that is too weak for efficient O$_2$ activation in acidic media, leading to high activation overpotentials \cite{orr_norskov_2018}. However, this weak binding can become advantageous in alkaline electrolytes, where the different reaction environment and mechanism shift the optimal adsorption strength toward weaker binding. Literature reports confirm that Ag-Au alloys are both stable and highly active under alkaline conditions \cite{ORR_AuAg_2024, ORR_AuAg_2025}. Consequently, the AgAu(110) surface, with its +0.14 eV adsorption energy shift, is identified as a particularly promising candidate for alkaline ORR catalysts. 

\section{Discussion}

Constructing universal MLIP models represents one of the most promising applications of machine learning within chemistry and materials science. However, developing these MLIP models typically requires extensive high-quality datasets and substantial computational resources for training. Additionally, their performance is inherently limited by insufficient representation of atomic forces at regions of the PES far from local minima. This limitation arises because the training data predominantly comprises structures near equilibrium, leading to inaccurate model predictions during the optimisation of structures distant from these equilibrium regions. In this study, we demonstrate that the DBCata model effectively learns the transition directly between initial and relaxed adsorption structures, circumventing the need to approximate the entire PES. As a result, DBCata could robustly sample local minima, using initial structures that may be far from equilibrium, a regime in which conventional MLIPs often struggle. Taken together, the combination of DBCata and MLIPs enables efficient acceleration of accurate adsorption energy calculation and catalyst computational screening. 

This work demonstrates the potential of generative models for predicting adsorption structures, thereby accelerating catalyst design. Nevertheless, the current DBCata mapping is learned from pairwise distributions between initial and relaxed structures. Consequently, initial structures must be carefully initialised to fall within the training distribution, following a fixed rule. It may limit the extension of DBCata to complex heterogeneous catalysis. It is possible to integrate stochastic optimal control with diffusion bridge or bridge matching methods to reduce reliance on carefully curated initial-relaxed pairs by learning the adsorption process directly from PES gradients in the future, thereby circumventing the computationally expensive DFT relaxations currently required for dataset generation and improving generalisation. Combining these advances with DBCata could substantially enhance its applicability in more catalyst systems.

\section{Methods}

\subsection{DBCata model architecture} \label{DBCata Method}

\subsubsection{Generative model for accurate structure prediction}  

The DBCata model is based on the Brownian bridge approach \cite{BBDM_Li_2023}, which is a stochastic process that connects two probability distributions. Specifically, a Brownian bridge process with a fixed starting point $\mathbf{x}_0$ and endpoint $\mathbf{x}_T$ can be formulated as a continuous-time stochastic process $\mathbf{x}_t$ satisfying the following probability distribution:
\begin{equation}
    \mathbf{x}_t \sim \mathcal{N}(\mathbf{x}_0 + \frac{t}{T}(\mathbf{x}_T - \mathbf{x}_0), t(1 - \frac{t}{T})\mathbf{I}).
\end{equation}

Using $\mathbf{x}$, $\mathbf{y}$ to denote the coordinates of the relaxed and initial structures, respectively, we can define the forward diffusion process as follows:
\begin{equation}
    q(\mathbf{x}_t|\mathbf{x}_0,\mathbf{y})=\mathcal{N}\left(\mathbf{x}_t;(1 - m_t)\mathbf{x}_0 + m_t\mathbf{y},\delta_t \mathbf{I}\right), \quad \mathbf{x}_0 = \mathbf{x}, \quad m_t=\frac{t}{T}
\end{equation}
where $\delta_t$ is the variance of the noise at time $t$. In practice, $\delta_t$ is set to $2s(m_t - m_t^2)$ with $s$ being a hyperparameter. The forward diffusion process is a Markov process, and the transition kernel can be expressed as: 
\begin{equation}
    q(\mathbf{x}_t | \mathbf{x}_{t-1}, \mathbf{y}) = \mathcal{N}\left(\mathbf{x}_t; \frac{1 - m_t}{1 - m_{t-1}} \mathbf{x}_{t-1} + (m_t - \frac{1-m_t}{1-m_{t-1}}m_{t-1})\mathbf{y}, \delta_{t|t-1}\mathbf{I}\right)
\end{equation}
The reverse process aims to predict $\mathbf{x}_{t-1}$ based on $\mathbf{x}_t$:
\begin{equation}
    p_{\theta}(\mathbf{x}_{t-1} | \mathbf{x}_t, \mathbf{y}) = \mathcal{N}\left( \mathbf{x}_{t-1}; \boldsymbol{\mu}_{\theta}(\mathbf{x}_t, t), \delta_t\mathbf{I} \right)
\end{equation}
where $\boldsymbol{\mu}_{\theta}$ can be obtained from neural network prediction. The training objective is to maximise ELBO:
\begin{equation}
    \begin{split}
    ELBO = &-\mathbb{E}_q\bigl(D_{KL}(q(\mathbf{x}_T|\mathbf{x}_0,\mathbf{y})||p(\mathbf{x}_T|\mathbf{y})) \\
    &+\sum_{t=2}^{T}D_{KL}(q(\mathbf{x}_{t-1}|\mathbf{x}_t,\mathbf{x}_0,\mathbf{y})||p_{\theta}(\mathbf{x}_{t-1}|\mathbf{x}_t,\mathbf{y})) \\
    &- \log p_{\theta}(\mathbf{x}_0|\mathbf{x}_1,\mathbf{y})\bigr)
    \end{split}
\end{equation}
Empirically \cite{ddpm_ho_2020, ddim_song_2020, sgm_song_2020}, the reparameterisation trick is used, and the training objective can be simplified as:
\begin{equation}
    \mathcal{L} = \left\| m_t(\mathbf{y} - \mathbf{x}) + \sqrt{\delta_t}\boldsymbol{\epsilon} - \boldsymbol{\epsilon}_{\theta}(\mathbf{x}_t, t) \right\|
\end{equation}
where $\boldsymbol{\epsilon}_{\theta}$ is the objective predicted by the neural network directly. And then, the model can be trained efficiently by minimising the simplified loss function.

\subsubsection{Multi-graph and nearest atom interpolation method for periodic boundary conditions} 

Unlike the original Brownian bridge model, DBCata is designed to work with periodic boundary conditions, which are essential for modelling adsorption structures on surfaces. To achieve this, we utilise a multi-graph approach to identify each atom's neighbours within a cutoff radius across multiple periodic images, and then the distance between atom $j$ and atom $i$ is determined as:
\begin{equation}
    d_{ij} = \left\| \mathbf{r}_i - \mathbf{r}_j + k_1\mathbf{L}_x + k_2\mathbf{L}_y \right\| < R_c
\end{equation}
where $\mathbf{L}$ is the lattice vector shift (only the x and y directions are considered because of the vacuum space in the z direction). $k$ can take values of -1, 0, or 1, and $R_c$ is the cutoff radius. 

Besides, we introduce a nearest-atom interpolation method that ensures the transition probabilities during the Brownian bridge process inherently satisfy periodic boundary conditions. This method is implemented by modifying the linear interpolation function and the transition kernel. First, we modify the mean value $\mathbf{x}_t$ in equation (2) by replacing the original linear interpolation with a nearest-atom interpolation:
\begin{equation}
    \mathbf{x}_t = \mathbf{x}_0 + m_t (\mathbf{L}\cdot\text{round}\left(\mathbf{L}^{-1}(\mathbf{y} - \mathbf{x}_0)\right))
\end{equation}
where $\text{round}(\cdot)$ is the rounding function. Accordingly, the loss function of DBCata in equation (6) is modified as follows:
\begin{equation}
    \mathcal{L} = \left\|  m_t (\mathbf{L}\cdot\text{round}\left(\mathbf{L}^{-1}(\mathbf{y} - \mathbf{x})\right)) + \sqrt{\delta_t}\boldsymbol{\epsilon} - \boldsymbol{\epsilon}_{\theta}(\mathbf{x}_t, t)  \right\|
\end{equation}
by adding the nearest atom interpolation term. This modification circumvents long-displacement interpolations when atoms are close to the periodic boundary, which can lead to large errors in the predicted structures.

Second, we modify the transition kernel in equation (4) by reparameterising the predicted $\boldsymbol{\mu}_{\theta}(\mathbf{x}_t, t)$ as in equation (11), 
\begin{equation}
    \boldsymbol{\mu}_{\theta}(\mathbf{x}_t, t) = (1 - m_{t-1})\mathbf{x} + m_{t-1}\mathbf{y} + \sqrt{\frac{\delta_{t-1} - \sigma_{t}^2}{\delta_t}}(\mathbf{x}_t - (1-m_t)\mathbf{x}-m_t\mathbf{y}) 
\end{equation}
\begin{equation}
    \tilde{\boldsymbol{\mu}}_{\theta}(\mathbf{x}_t, t) = \mathbf{x} + m_{t-1} \tilde{\mathbf{r}} + \sqrt{\frac{\delta_{t-1} - \sigma_{t}^2}{\delta_t}}(\mathbf{L}\cdot\text{round}\left(\mathbf{L}^{-1}(\mathbf{x}_t - (\mathbf{x} + m_t\tilde{\mathbf{r}}))\right)) 
\end{equation}
\begin{equation}
\tilde{\mathbf{r}} = \mathbf{L}\cdot\text{round}\left(\mathbf{L}^{-1}(\mathbf{y} - \mathbf{x})\right)
\end{equation}
where the nearest-atom interpolation is also used to ensure all the states of atom coordinates are in the same periodic image cell. And finally, all the predicted coordinates at the time of $t$ are scaled to be inside the original cell by:
\begin{equation}
    \mathbf{x}_t = \mathbf{L}\cdot\text{round}\left(\mathbf{L}^{-1}(\mathbf{x}_t)\right)
\end{equation}

\subsubsection{Graph neural network for equivariant representation learning}

As shown in Fig.\ref{overview}b, the DBCata model employs an equivariant graph neural network to learn the transition kernel. Here, in adsorption or surface structure representation, equivariance actually means that the model output should be invariant to the permutation of atoms and the periodic translation of the entire system, e.g., periodic translation invariant. To achieve this, we use the PaiNN architecture \cite{PaiNN_Schutt_2021} as the transition kernel. This model is based on the message passing framework \cite{mpnn_gilmer_2017}, which is a common approach in graph neural networks to learn equivariant representations of graph-structured data. A general interaction layer of a MPNN block with a time conditional embedding $t$ can be expressed as:
\begin{equation}
m_i^{l+1} = \sum_{j\in \mathcal{N}(i)} \mathbf{M}_l\bigl(h_i^l(t),\;h_j^l(t),\;\vec r_{ij}\bigr)
\end{equation}
\begin{equation}
h_i^{l+1}(t) = \mathbf{U}_l\bigl(h_i^l(t),\;m_i^{l+1}\bigr)
\end{equation}
where $h_i^l$ and $m_{ij}^l$ are the node and message feature vectors at layer $l$. $\mathbf{M}_l$ and $\mathbf{U}_l$ are the message and update functions, respectively. Before the MPNN block, fractional coordinates of atoms $F$ are used to ensure that all the atoms are in the same periodic image cell. And then, the neighbours of each atom can be identified by the cutoff radius $R_c$. We use atom embedding and the Gaussian radial basis function (RBF) \cite{rbf_fornberg_2011} to encode the atom features and distance features. Additionally, a periodic positional encoding is used by adding the Fourier features of the fractional distances. Ablation studies of the MPNN architecture are provided in Supplementary Table 4.

In this work, each MPNN block layer consists of surface atomic interactions, all the atomic interactions, and an update function. After multiple MPNN blocks, the final output is obtained by a readout function.

\subsection{Details and configurations for model training}

\subsubsection{DBCata training}

At the training stage, the total number of diffusion steps $T$ was set to 100, and the maximum noise variance $\delta_{max}$ was set to 0.05 to ensure that the Brownian bridge process sufficiently covers the entire path of structure relaxation. While at the validation or generation stage, we adopted a deterministic sampling method based on ODE to guarantee that the generated structures are invariant to repeated samplings. The total number of sampling steps was set to 20, and the variance of noise is zero. The cutoff radius $R_c$ was set to 4.0 \text{\AA}. Atoms below the surface are masked to be fixed during the diffusion process. The hidden dimension of the graph neural network was set to 256, and the number of layers was set to 4. The learning rate was set to $10^{-4}$ and an exponential decay learning rate scheduler was used. AdamW optimiser \cite{adam_kingma_2015, adamw_loshchilov_2017} was used without weight decay. The training process takes 1000 epochs to converge with a batch size of 64. All the complete related hyperparameters are provided in Supplementary Table 3.

\subsubsection{Outlier detection}

To get labelled outlier data, we used initial structures from the Catalysis-Hub dataset and generated predicted relaxed structures by the pretrained DBCata model at different noise levels. The noise level coefficients were set to 0.0, 0.5, and 1.0. Then we calculated the DMAE values of all 265197 generated structures and labelled them as outliers if the DMAE value is larger than 0.05 \text{\AA}. Additionally, we used chemical heuristics to label the structures as outliers with four such scenarios: (1) Collision of atoms: if the distance between two atoms is less than 0.8 plus the sum of their covalent radii. (2) Dissociation of adsorbate: if the adsorbate is dissociated into two or more parts. (3) Desorption of adsorbate: if the adsorbate is desorbed from the surface. (4) Surface reconstruction: if the surface atoms are reconstructed, it is no longer comparable to the corresponding clean surface. Most functions for heuristic detection were adapted from AdsorbML code \cite{AdsorbML_Ulissi_2023}. 

The labelled data were then split into training and test sets in a ratio of 8:2. The classifier model was the same architecture as the graph neural network in DBCata, but with scalar output as the confidence score. The training batch size was set to 256, and the training epoch was set to 20. The learning rate was set to a constant $10^{-3}$. Weight balancing was used to balance the positive and negative samples. 

\subsubsection{Initial guess structure generation} \label{IS}

To ensure that the inputs to DBCata follow the same distribution, adsorption configurations were generated using an automated workflow following the procedure of the Catalysis-Hub dataset \cite{CatHub_Thomas_2019}. The screening binary alloy slabs  were automatically constructed by CatKit. The optimised lattice constants of the corresponding bulk were sourced from Catalysis-Hub and instantiated bulk crystals with ASE \cite{ase_Larsen_2017}.Each slab consisted of three atomic layers, with the bottom two layers fixed during subsequent structural relaxations. A vacuum spacing of 10 \text{\AA} was applied along the $z$ direction, and primitive surface cells with trimmed-layer construction were employed while preserving the alloy ordering of the input structures. We enumerated all non-equivalent surface terminations and generated one slab per termination. The resulting slab set was collected for downstream adsorption placement and DBCata-driven relaxation. The adsorption configurations at top, bridge, and hollow sites on the topmost surface layer were constructed via Delaunay triangulation in the surface plane and subsequently reduced through distance-based clustering and symmetry analysis to yield a unique set of sites. Each site was then shifted by 1.5 \text{\AA} along the surface normal, and the adsorbate molecule was translated such that its designated adsorption atom coincided with the origin and, where applicable, rotated to align its molecular z-axis with the surface normal before being placed on the slab.

\subsection{Evaluation indicators}

To evaluate the accuracy of the structures generated by our model, we employed DMAE (Distance Mean Absolute Error), which quantifies the discrepancies between the generated structures and those relaxed using DFT. DMAE is defined as:
\begin{equation}
    \text{DMAE} = \frac{1}{N^2} \sum_{i=1}^{N} \sum_{\substack{j=1}}^{N} \left| d_{\text{DFT}, i, j} - d_{\text{DBCata}, i, j} \right|
\end{equation}
where N denotes the total number of atoms in the system, $d_{\text{DFT}, i, j}$ and $d_{\text{DBCata}, i, j}$ are the distances between atom $i$ and atom $j$ in the DFT-relaxed and DBCata-generated structures, respectively. This distance between atoms is determined as the shortest distance across all periodic images, thereby inherently satisfying the periodic boundary conditions and ensuring the validity of the evaluation.

The success ratio of prediction accuracy was used for evaluating the DFT calculation results of generated structures compared to DFT-relaxed structures, which is defined as:
\begin{equation}
    \eta = \frac{1}{m}\sum_{i=1}^{m} \mathbb{I}\!\left(|E_i^{\mathrm{DBCata}} - E_i^{\mathrm{DFT}}| \leq \epsilon\right),
\end{equation}
where $m$ is the total number of structures. $\epsilon$ is the energy threshold. $\mathbb{I}(\cdot)$ denotes the indicator function, which takes the value 1 if the condition is satisfied and 0 otherwise.

\subsection{Computational details of DFT and MLIP} \label{Computational details}

All spin-polarised DFT calculations were performed using Vienna Ab initio Simulation Package (VASP) \cite{vasp1, vasp2, vasp3}, employing the RPBE exchange correlation functional. The single-point calculations were performed with a plane wave energy cutoff of 450 eV, and the convergence criteria for energy were set to $10^{-5}$ eV. Geometry relaxations were performed with convergence criteria of $10^{-5}$ eV for energy and 0.05 eV/\text{\AA} for forces. For slab calculations, Brillouin-zone sampling was employed using a Monkhorst-Pack k-point grid \cite{monkhorst_1976} chosen such that the product of the number of k-points $N_k$ and the corresponding supercell length $L$ in each periodic direction was approximately 30 \text{\AA}. For isolated molecular calculations, the $\Gamma$-point \cite{gamma_1973} was used.

Adsorption energies were calculated using the isolated gas-phase adsorbate as the reference. 
For a generic adsorbate $X$ (e.g., $X = \mathrm{C_aH_bO_cN_dS_e}$), the adsorption energy is defined as
\begin{equation}
    \Delta E_{\mathrm{ads}}(X)
    = E_{X^*} - E_{\mathrm{slab}} - E_{X(\mathrm{g})},
\end{equation}
where $E_X$ denotes the DFT total energy of species $X$, $E_{\mathrm{slab}}$ is the energy of the clean slab, and the superscript $^*$ indicates an adsorbed species. For instance, to verify the accuracy of DBCata by fitting scaling relationships and to identify promising candidates, we computed the adsorption energies of O and OH species as
\begin{equation}
    \Delta E_{\mathrm{ads}}(\mathrm{O}) 
    = E_{\mathrm{O^*}} - E_{\mathrm{slab}} - E_{\mathrm{O(g)}},
\end{equation}
\begin{equation}
    \Delta E_{\mathrm{ads}}(\mathrm{OH}) 
    = E_{\mathrm{OH^*}} - E_{\mathrm{slab}} - E_{\mathrm{OH(g)}}.
\end{equation}

For MLIP geometry relaxation, we employed the UMA-OC20 pretrained model, which is a mixture of experts (MoE) architecture-based graph neural network model trained on large datasets, including OC20-S2EF. Outputs of UMA-OC20 are the energy and forces of the input system. The L-BFGS algorithm \cite{LBFGS_Liu_1989} implemented in ASE packages \cite{ase_Larsen_2017} was used for geometry optimisation. The convergence criterion of the force was set to 0.05 eV/\text{\AA} and the maximum step of relaxation was 500.

\section*{Data and code availability}

The code and datasets generated during and/or analysed during the current study will be made publicly available upon the formal publication of this manuscript.


\bibliography{sn-bibliography}

\section*{Acknowledgements}
This work was supported by the National Key Research and Development Program of China (2023YFA1507601 and 2021YFA1500700). We thank the Centre for High Performance Computing at Shanghai Jiao Tong University for providing the computing resources of the Siyuan-1 cluster.

\section*{Author information}
\subsection*{Author contributions}
S.Huo wrote the code, performed theoretical simulations, and wrote the paper. X.-M.C. designed the study, analysed the data, and wrote the paper. Both authors discussed the results and improved the manuscript.

\subsection*{Corresponding authors}
Correspondence to Xiao-Ming Cao.

\section*{Ethics declarations}
\subsection*{Competing Interests}
All authors declare no competing interests.

\section*{Supplementary information}
Supplementary Figs. 1-4 and Tables 1-5.

\newpage

\newpage
\appendix 
\setcounter{page}{1} 
\renewcommand{\thepage}{S\arabic{page}} 

\setcounter{figure}{0}
\renewcommand{\thefigure}{S\arabic{figure}}
\setcounter{table}{0}
\renewcommand{\thetable}{S\arabic{table}}
\setcounter{equation}{0}
\renewcommand{\theequation}{S\arabic{equation}}

\begin{center}
    \vspace*{2.0cm}
    
    {\Large \bfseries Supplementary Information for: \\[0.5em]
    Accelerating High-Throughput Catalyst Screening by Direct Generation of Equilibrium Adsorption Structures}
    
    \vspace{1.0cm}
    
    {\large Songze Huo$^{1}$ and Xiao-Ming Cao$^{1,*}$}
    
    \vspace{0.5cm}
    
    {\small \textit{$^{1}$State Key Laboratory of Synergistic Chem-Bio Synthesis, School of Chemistry and Chemical Engineering, Shanghai Jiao Tong University, 800 Dongchuan Road, Shanghai, 200240, Shanghai, China}}
    
    \vspace{0.5cm}
    
    {\small *Email: xmcao@sjtu.edu.cn}
    
\end{center}
\vspace{1.0cm}

\newpage 

\subsection*{1 Supplementary tables}

\begin{center}
  \textbf{Table S1. Time comparison of DBCata with different timesteps, UMA-OC20 optimisation and DFT relaxation}
  \vspace{0.2cm} 
  
  \begin{tabular}{ccccc}
    \hline
    Method & \makecell{Time (s) \\ per structure} & Speedup & \makecell{Batch Parallel \\ (1 for non-parallel)} & DMAE \\
    \hline
    DFT & 851.284 & 1x & 1 & / \\
    UMA-OC20 & 9.777 & 87x & 1 & 0.1133 \\
    DBCata(100) & 1.039 & 819x & 1 & 0.0349 \\
    DBCata(20) & 0.212 & 4015x & 1 & 0.0353 \\
    DBCata(100) & 0.009 & 96736x & 256 & 0.0349 \\
    DBCata(20) & 0.002 & 425640x & 256 & 0.0353 \\
    \hline
  \end{tabular}
\end{center}
\vspace{\baselineskip}
Tests were conducted on the Catalysis-Hub test dataset, comprising a $p(2 \times 2) \times 3$-layers slab and the adsorption species of CH$_x$, OH$_x$, NH$_x$, and SH$_x$. 

DBCata(n) denotes the DBCata model with n sampling steps. The speedup was calculated by comparing the computational time of each model with that of DFT relaxation. DMAE refers to the distance mean absolute error between model-relaxed and DFT-relaxed structures. UMA and DBCata evaluations were performed on a single NVIDIA GeForce RTX 4090 GPU with 24 GB memory, while DFT relaxations were carried out using the RPBE functional with a plane-wave energy cutoff of 450 eV.

\newpage

\begin{center}
  \textbf{Table S2. Analysis of adsorption species in the Catalysis-Hub dataset}
  \vspace{0.2cm}

  \begin{tabular}{cccc}
    \hline
    Adsorbate & Count (Train / Val) & Percentage (Train / Val) & DMAE (Val)\\
    \hline
    H & (9164 / 2252) & (15.55\%, 15.29\%) & 0.0268 \\
    C & (8598 / 2077) & (14.59\%, 14.10\%) & 0.0410 \\
    CH & (1862 / 492) & (3.16\%, 3.34\%) & 0.0317 \\
    CH$_2$ & (1763 / 473) & (2.99\%, 3.21\%) & 0.0499 \\
    CH$_3$ & (1732 / 438) & (2.94\%, 2.97\%) & 0.0533 \\
    O & (8747 / 2231) & (14.84\%, 15.15\%) & 0.0333  \\
    OH & (4582 / 1129) & (7.77\%, 7.67\%) & 0.0285 \\
    H$_2$O & (1757 / 435) & (2.98\%, 2.95\%) & 0.0667 \\
    N & (8652 / 2199) & (14.68\%, 14.93\%) & 0.0350 \\
    NH & (1836 / 490) & (3.11\%, 3.33\%) & 0.0325 \\
    S & (8398 / 2051) & (14.25\%, 13.93\%) & 0.0274 \\
    SH & (1851 / 461) & (3.14\%, 3.13\%) & 0.0473 \\
    \hline
  \end{tabular}
\end{center}

\newpage

\begin{center}
    \textbf{Table S3. Main hyperparameters of DBCata model}
\end{center}

Training hyperparameters.
\begin{center}
  \begin{tabular}{lll}
    \hline
    Parameter & Value & Notes \\
    \hline
    debugging & \texttt{False} &  \\
    flow & \texttt{bbdm} & options: \texttt{bbdm}, \texttt{rf} \\
    coord & \texttt{cartesian} & options: \texttt{cartesian}, \texttt{fractional} \\
    epoch & \texttt{2000} & (alt: \texttt{4000}) \\
    model\_name & \texttt{adspainn} & options: \texttt{painn}, \texttt{egnn}, \texttt{adspainn}, \texttt{equiformerv2} \\
    batch\_size & \texttt{64} & (alt: \texttt{128}; e.g., grad accum \(\times 4\)) \\
    lr & \texttt{1e-4} & (alt: \texttt{1e-3}) \\
    schedule\_gamma & \texttt{0.999} &  \\
    num\_workers & \texttt{0} &  \\
    matmul\_precision & \texttt{highest} &  \\
    clip\_grad & \texttt{True} &  \\
    loss\_type & \texttt{l1} & \texttt{l1} for \texttt{bbdm}, \texttt{l2} for \texttt{rf} \\
    fixed & \texttt{True} & fix atoms below \\
    train\_objective & \texttt{grad} & options: \texttt{grad}, \texttt{noise}, \texttt{ysubx} (bbdm only) \\
    frac\_noise & \texttt{True} & auto \texttt{False} when \texttt{cartesian} \\
    write\_outputs & \texttt{False} &  \\
  \end{tabular}
\end{center}

\par\bigskip

Ads-PaiNN hyperparameters.
\begin{center}
  \begin{tabular}{lll}
    \hline
    Parameter & Value & Notes \\
    \hline
    cutoff & \texttt{4.0} &  \\
    hidden\_channels & \texttt{256} &  \\
    out\_channels & \texttt{3} &  \\
    num\_rbf & \texttt{256} &  \\
    rbf & \texttt{gaussian} &  \\
    envelope & \texttt{polynomial (exp=5)} &  \\
    num\_layers & \texttt{4} &  \\
    n\_frequencies & \texttt{40} & Fourier features \\
    scalar & \texttt{False} & use scalar \texttt{xh\_out} or not \\
    ftbasis & \texttt{True} & (alt: \texttt{True}) use Fourier basis or not \\
    \hline
  \end{tabular}
\end{center}

\par\bigskip

Diffusion hyperparameters.
\begin{center}
  \begin{tabular}{lll}
    \hline
    Parameter & Value & Notes \\
    \hline
    sample\_per\_epoch & \texttt{5} &  \\
    num\_timesteps & \texttt{100} &  \\
    sample\_mt\_mode & \texttt{cosine} & options: \texttt{linear}, \texttt{sin}, \texttt{cosine} \\
    max\_var & \texttt{0.05} & range: 0.001\,--\,0.05 \\
    sample\_steps & \texttt{20} &  \\
    skip\_sample & \texttt{True} &  \\
    sample\_mode & \texttt{linear} & options: \texttt{linear}, \texttt{cosine}, \texttt{all} \\
    eta & \texttt{0.0} &  \\
    \hline
  \end{tabular}
\end{center}

\newpage

\begin{center}
  \textbf{Table S4. Ablation experiments of DBCata on Catalysis-Hub test dataset}
  \vspace{0.2cm}

  \small
  \begin{tabular}{l c c c c c c}
  \toprule
  Variant & \makecell{Fractional\\coordinate} & \makecell{Gaussian Basis\\Function} & \makecell{Fourier Basis\\Function} & \makecell{DMAE\\$\downarrow$}  \\
  \midrule
  Baseline    & \xmark & \cmark & \xmark & 0.0409  \\
      & \cmark & \xmark & \cmark & 0.0621 \\
   & \xmark & \cmark & \cmark & \textbf{0.0353} \\
  \bottomrule
  \end{tabular}
\end{center}

\newpage

\begin{center}
  \textbf{Table S5. Classification metrics of outlier detection model on test dataset}
  \vspace{0.2cm}

  \small
  \begin{tabular}{l c c c c c c}
      \toprule
      Class & Precision & Recall & F1-score & Support \\
      \midrule
      negative     & 0.66 & 0.85 & 0.74 & 8,997  \\
      positive     & 0.96 & 0.89 & 0.92 & 35,187 \\
      \midrule
      \multicolumn{1}{l}{Accuracy} & \multicolumn{3}{c}{0.88} & 44,184 \\
  \bottomrule
  \end{tabular}
\end{center}
\vspace{\baselineskip}
Negative: outlier; positive: normal structure.

\newpage

\subsection*{2 Supplementary figures}

\subsubsection*{Figure S1}

\vspace{0.2cm}

\begin{center}
  \includegraphics[width=0.7\textwidth]{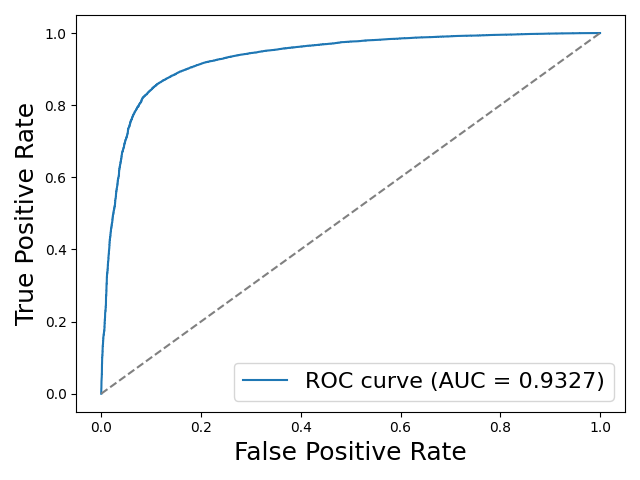}
  
  \textbf{ROC (Receiver Operating Characteristic) curve of outlier detection model}
\end{center}

\newpage

\subsubsection*{Figure S2}

\vspace{0.2cm}

\begin{center}
  \includegraphics[width=0.8\textwidth]{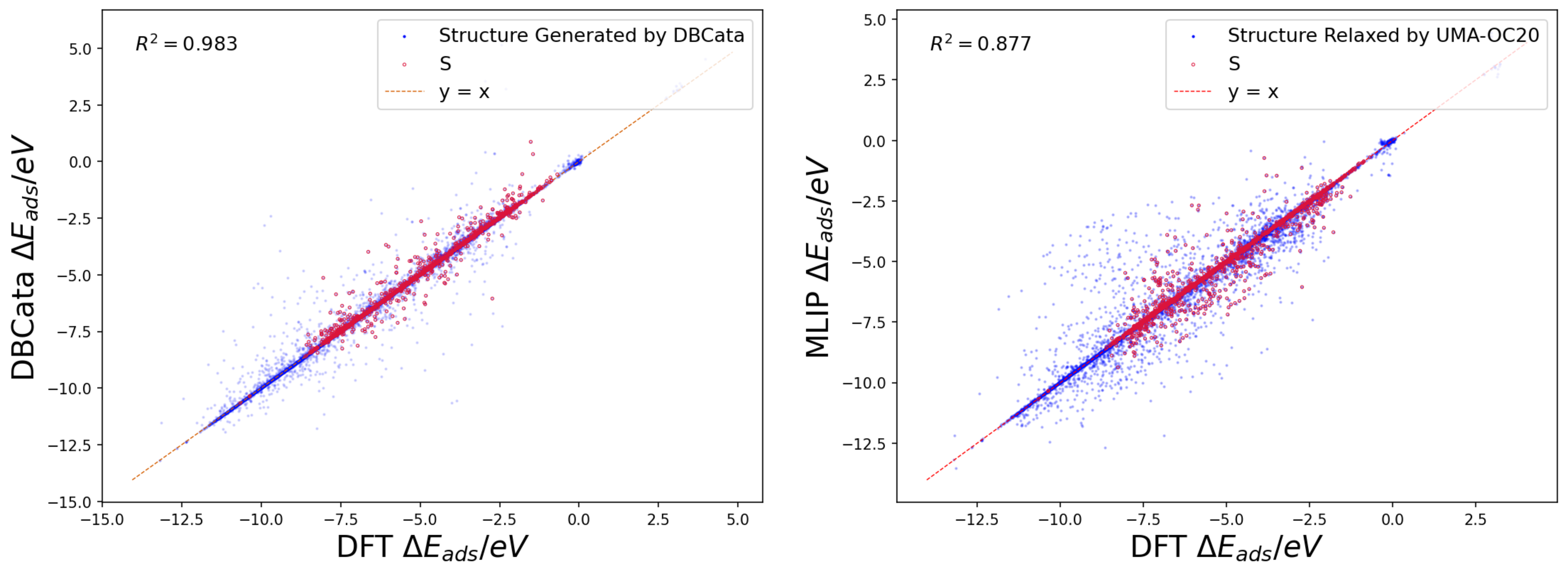}
  
  \textbf{Out-of-Distribution Evaluation: UMA-OC20 on Unseen Sulfur-Containing Adsorbates}
\end{center}
\vspace{\baselineskip}

Adsorption energies of systems from the Catalysis-Hub test dataset, including sulphur-containing adsorbates, which are highlighted in red circles. 

Since sulphur (S)-containing adsorbates were not included in the UMA training data, we isolated these species for a separate evaluation. As shown in Figure, after relaxation with UMA-OC20, they achieve prediction accuracy comparable to that of other adsorbates. We attribute this robustness to the exposure of UMA during training to a large dataset that nevertheless contains S-bearing structures, even though S-containing adsorbates themselves were not present.

\newpage 

\subsubsection*{Figure S3}

\vspace{0.2cm}

\begin{center}
  \includegraphics[width=0.8\textwidth]{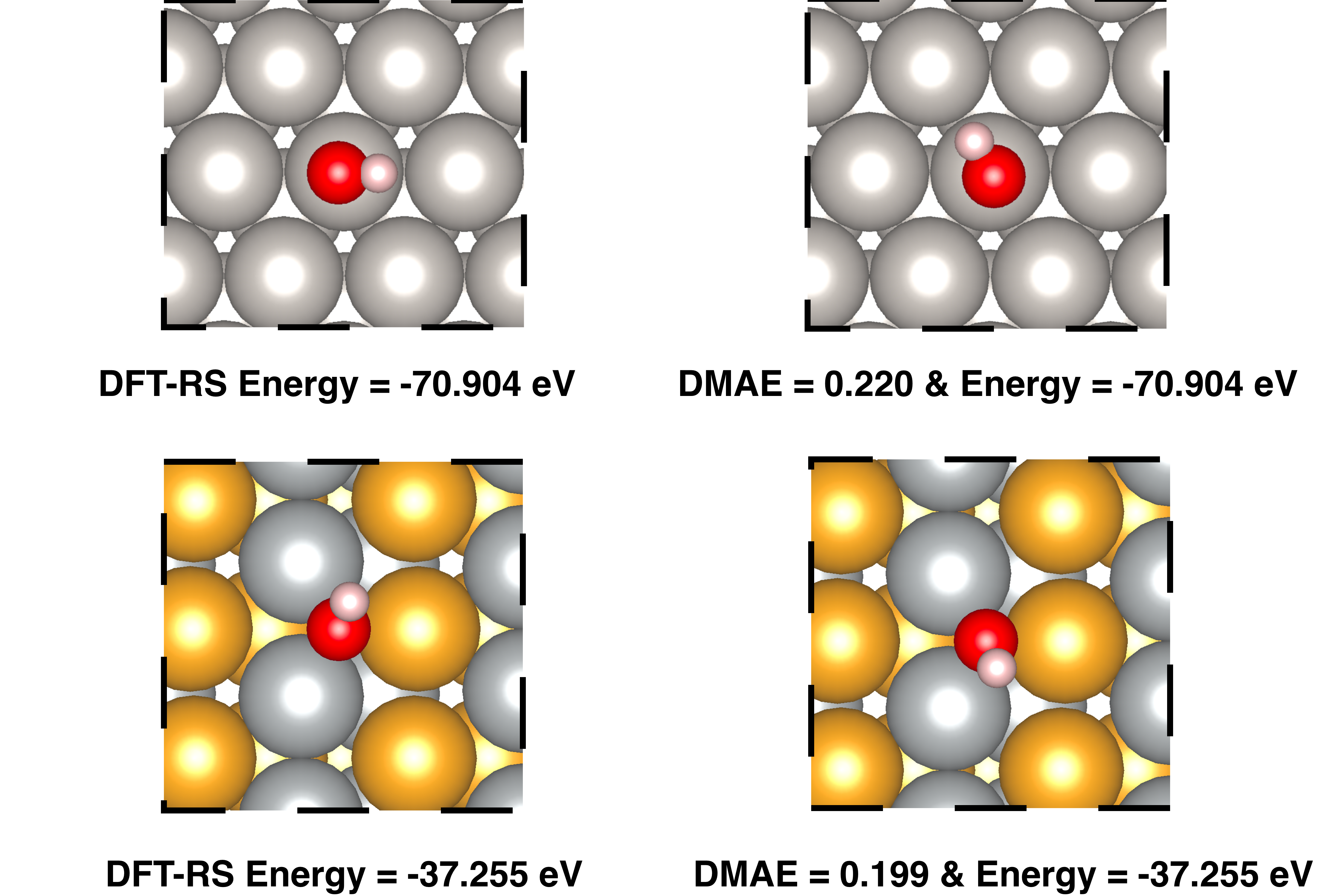}

  \textbf{Illustration of spuriously elevated DMAE values caused by distinct O-H bond orientations in energetically equivalent adsorption structures.}
\end{center}
\vspace{\baselineskip}

Adsorption of OH on Pt(111) and AuAg(111) surfaces. The DBCata-generated structures (right) are energetically identical to the DFT-relaxed structures (left), while exhibiting a structural DMAE of $0.220\,\text{\AA}$ and $0.199\,\text{\AA}$, respectively.

\newpage 

\subsubsection*{Figure S4}

\vspace{0.2cm}

\begin{center}
  \includegraphics[width=0.8\textwidth]{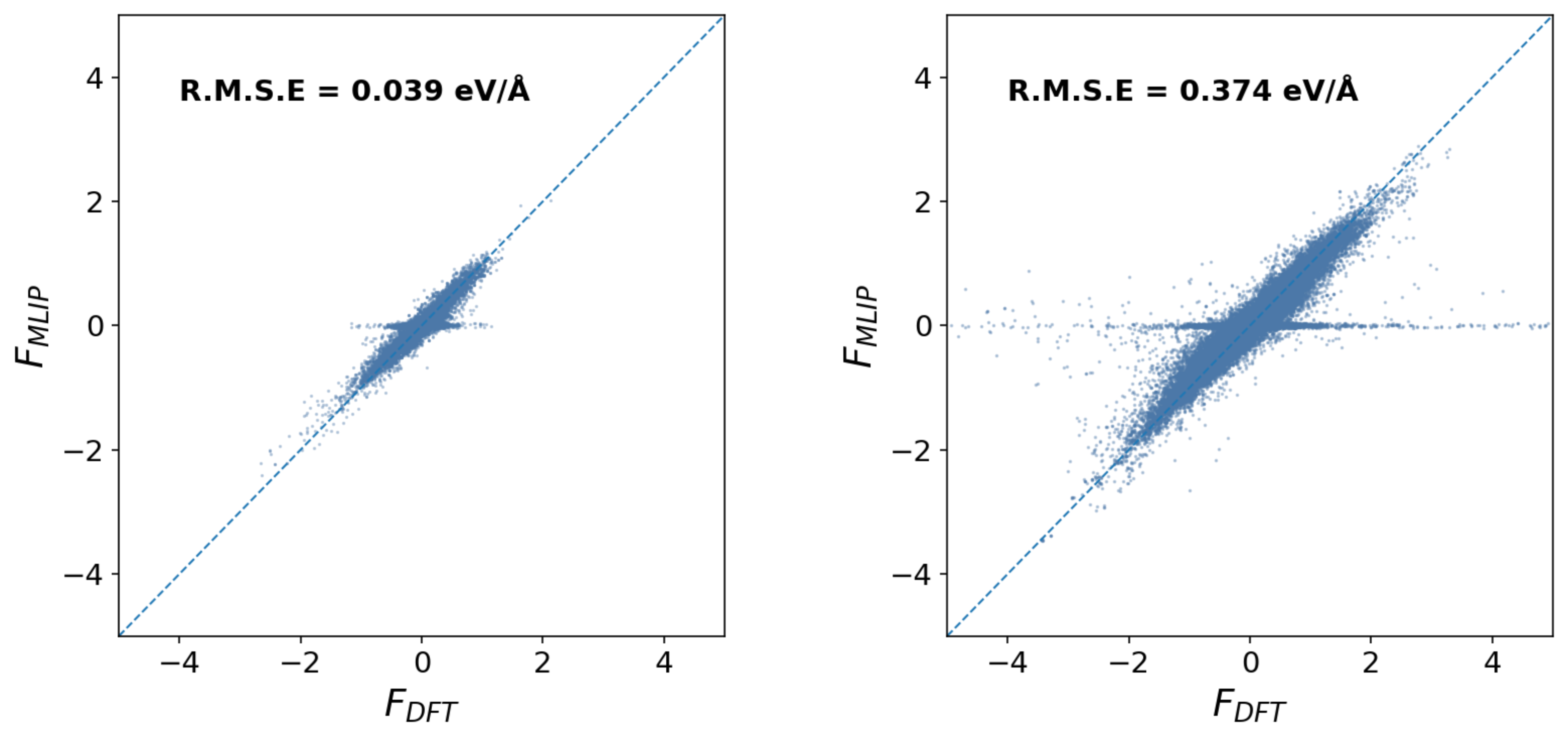}

  \textbf{Evaluation of atomic forces predicted by UMA-OC20 against the RPBE functional.}
\end{center}
\vspace{\baselineskip}

(Left) Parity plot of atomic forces for 5,081 near-equilibrium structures. (Right) Parity plot for 9,647 far-from-equilibrium structures. The right panel exhibits a larger fraction of points deviating from the $y=x$ parity line. The evaluation encompasses all atoms, including those subjected to positional constraints (fixed) during relaxation.

\end{document}